# Heterogeneous AAV Logistics Task Allocation: A Reinforcement Learning Enhanced Overlapping Coalition Formation Game Approach

Yuze Zhou, Jingliang Sun, Junzhi Li, Jianxin Zhong, Zihan Wang, and Teng Long

*Abstract*—**In dynamic urban logistics, the stochastic emergence of time-sensitive tasks poses a significant optimality challenge for heterogeneous AAVs logistics task allocation. To address this problem, a reinforcement learning enhanced overlapping coalition formation game approach is proposed. A dynamic task allocation model is established, where global optimality is mathematically quantified by a generalized logistics cost coupling service quality and resource consumption. To deal with the time-varying task sets induced by stochastic order arrivals, a transformer-based soft actor-critic network is designed. By leveraging multi-head self-attention to encode variable-length logistics states and capture task-wise spatiotemporal dependencies, the learned policy adaptively guides coalition updates, replacing heuristic rules in the overlapping coalition formation game. On this basis, heterogeneous AAVs can form more efficient overlapping coalitions for dynamic logistics tasks. The resulting coalition formation process is proven to constitute an exact potential game, which guarantees convergence to a Nash-stable equilibrium within a finite number of iterations. Numerical simulations demonstrate that the proposed algorithm effectively improves the optimality of task allocation under the generalized logistics cost criterion. In a scenario with 32 AAVs and 80 tasks, our algorithm achieves a 39.76% cost reduction compared with the heuristic OCF baseline. Indoor flight experiments further validate its practicality.**

## I. Introduction

In recent years, autonomous aerial vehicles (AAVs) have emerged as a cornerstone of the low-altitude economy, facilitating applications such as logistics delivery and environmental monitoring, which significantly improve civilian life [1], [2]. Specifically in urban logistics, effective collaboration among heterogeneous AAVs can substantially enhance delivery efficiency and reduce transportation costs. Task allocation, as a key technology for multi-AAV collaborative logistics, is essential for improving the global optimality of these dynamic logistics scenarios [3].

The authors are with Beijing Institute of Technology, Beijing 100081, China; and with the Key Laboratory of Dynamics and Control of Flight Vehicle, Ministry of Education, Beijing 100081, China; and with the Beijing Institute of Technology Chongqing Innovation Center, Chongqing, 401121, China; and also with the National Key Laboratory of Land and Air-based Information Perception and Control, Beijing 100081, China. (*Corresponding author: Teng Long*.)

Many representative works have been proposed to address the challenge of logistics task allocation [4], [5]. Existing studies can be classified into optimization-based methods [6], [7], market-based approaches [8], [9], and game-theoretic methods [10], [11]. Optimization-based methods typically formulate logistics task allocation as a centralized scheduling problem under constraints such as delivery time windows and payload limits [6], [7]. Although these methods can achieve high-quality solutions, their reliance on global information and centralized computation limits the scalability in dynamic logistics environments. Market-based approaches improve allocation flexibility by enabling AAVs to bid for tasks in a distributed manner [8], [9]. However, their performance depends heavily on the design of the negotiation mechanism and may degrade when logistics demands change rapidly. These limitations have motivated the adoption of game-theoretic methods, which provide a natural framework for modelling strategic interactions and stable cooperation in heterogeneous AAV logistics task allocation [10], [11].

Among game-theoretic methods, coalition formation game (CFG) is suitable for logistics systems, where AAVs often need to cooperate by pooling complementary resources such as payload, endurance, and route accessibility. In urban logistics scenarios, some tasks cannot be efficiently performed by a single AAV and rather require multiple AAVs to coordinate in order to meet delivery efficiency and resource constraints [12], [13]. CFG provides an effective framework for modelling such cooperation, as it enables AAVs to self-organize into utility-driven groups and form stable collaborative structures. However, in CFG-based task allocation methods, each AAV can participate in at most one coalition and serve only one task at a time. While this design simplifies coalition management, it may underutilize resource-abundant AAVs and fail to capture the operational flexibility required in logistics scenarios.

To overcome the limitations of CFG-based task allocation, overlapping coalition formation (OCF) game has been introduced to multi-AAV logistics task allocation. Under the OCF game, an AAV can allocate resources to multiple coalitions, thereby improving cooperation flexibility and resource utilization. Existing OCF game-based studies typically construct overlapping coalitions through random exploration or heuristic rules [14], [15], [10]. For example, Li et al. [14] developed a heuristic OCF game allocation method that combines task-validity matching, ineffective-resource exit, and tabu lists to improve search efficiency under discrete resource constraints.

Qi et al. [15] proposed a sequential OCF game with a preference gravity-guided tabu search algorithm to guide coalition updates based on task-resource matching relations. Yan et al. [10] further improved scalability for large-scale AAV swarms by combining hierarchical clustering with random-exit-assisted coalition search. These methods enlarge the feasible coalition space and enhance the applicability of coalition-based task allocation in complex logistics environments. However, the increased flexibility of the OCF game also rapidly expands the decision space, making it difficult to efficiently obtain high-quality and stable allocation results as logistics tasks grow. Moreover, due to their heavy reliance on myopic, predefined rules, these methods often lack adaptability in dynamic logistics delivery and are prone to slow convergence and locally optimal coalition structures.

Fortunately, reinforcement learning (RL) provides a promising way to address these limitations, enabling AAVs to learn adaptive and forward-looking task-allocation policies through repeated interaction with dynamic logistics environments [16], [17], [18]. For example, Chen et al. [16] combined deep reinforcement learning with heuristic optimization to improve the joint scheduling quality of delivery AAVs, Gao et al. [17] designed a transfer-learning-based scheduling strategy for dynamic instant delivery that improves parcel throughput while reducing delivery cost, and Houran et al. [18] employed a policy-gradient-based attention model to generate efficient AAV delivery routes under capacity constraints. Compared with predefined coalition-update rules, RL offers the potential to evaluate current allocation schemes in view of their long-term impact on subsequent tasks, making it particularly appealing for OCF game-based logistics task allocation. However, integrating RL into the OCF game remains challenging. In dynamic logistics delivery, new orders may arrive continuously, causing the candidate task set to vary over time and making conventional fixed-size decision models less effective at capturing evolving task relationships and spatiotemporal dependencies [19], [20]. Consequently, conventional fixed-size RL architectures struggle to support effective overlapping coalition decisions in such environments.

Motivated by the above limitations, this paper develops a hybrid RL-driven OCF game algorithm for heterogeneous AAV logistics task allocation. Specifically, a forward-looking task-selection policy is learned first and then embedded into the OCF game to guide sequential overlapping coalition decisions in dynamic logistics environments. The main contributions are as follows.

1) A hybrid RL-guided OCF game algorithm is proposed for dynamic heterogeneous AAV logistics task allocation. A forward-looking task selection policy is pre-trained from dynamic logistics interactions through soft actor-critic policy evaluation and then embedded into the OCF game to replace conventional heuristic rules.

2) A Transformer-based policy network is designed to process variable-length task sequences in dynamic environments. By leveraging the multi-head self-attention mechanism to encode spatiotemporal dependencies, it enables AAVs to learn forward-looking policies.

3) The proposed OCF game is proven to be an exact potential game, which guarantees convergence to a Nash-stable equilibrium within a finite number of iterations.

## II. Problem Formulation

Consider a dynamic urban logistics scenario, as illustrated in Fig.1, where $N$ heterogeneous AAVs collaboratively execute $M$ randomly distributed tasks. The sets of AAVs, tasks, and depots are denoted as $\mathcal{N}=\{1,\dots n,\dots,N\}$, $\mathcal{M}=\{1,\dots,m,\dots,M\}$, and $\mathcal{D}=\{d_1,\dots,d_K\}$, respectively, where $K$ represents the number of depots.

In this logistics scenario, the total task set $\mathcal{M}$ is divided into pickup (P) tasks $\mathcal{M}_P$ and delivery (D) tasks $\mathcal{M}_D$, such that $\mathcal{M}=\mathcal{M}_P\cup\mathcal{M}_D$. Each task $m\in\mathcal{M}$ is characterized by a payload requirement $R_m$ and a strict time window (TW) $[T_m^{start},T_m^{end}]$. The logistics demand is considered divisible: a heavy-duty task where $R_m$ exceeds the capacity of a single AAV requires collaborative execution by AAV swarms, whereas light-duty tasks can be handled individually. Furthermore, to reflect the heterogeneity of the AAV swarm, each AAV $n$ is characterized by distinctive attributes: flight speed $v_n$, maximum payload capacity $B_n^{\max}$, and maximum flight range $D_n^{\max}$. All AAVs initiate their task sequences from the depots. During mission execution, the depots serve as on-demand resupply stations. AAVs can route to any depot in $\mathcal{D}$ to reset their payload states (either reloading or unloading), ensuring continuous operation through adaptable routing.

Unlike static allocation problems, the considered logistics scenario is dynamic. As illustrated by the state transition at $T=500$ s in Fig.1, new tasks with strict time windows stochastically emerge. AAVs must dynamically update their execution sequences to accommodate new logistics demands, thereby ensuring a timely and efficient response.

### *A. Problem Formulation*

The objective of the logistics scenario is to minimize the total system cost $\mathcal{J}$. Then, the problem is formulated as

$$\text{Problem 1:}\quad \min \mathcal{J}=\sum_{m\in\mathcal{M}} C_m(\mathcal{A}_m) \tag{1}$$

where $C_m(\mathcal{A}_m)$ represents the execution cost for task $m$, and the detailed formulation is given in Equation (4).

To address the problem, let $\mathcal{A}_m=\{a_m^{(1)},\dots,a_m^{(n)},\dots,a_m^{(N)}\}$ represent the payload allocation vector for task $m$, where $a_m^{(n)}\ge 0$ denotes the specific quantity of items to be delivered or picked up by AAV $n$. The set of AAVs collaboratively executing task $m$, referred to as the coalition, is defined as

$$\text{Coal}_m=\left\{n\in\mathcal{N}\,\middle|\,a_m^{(n)}>0\right\} \tag{2}$$

Specifically, task $m$ is regarded as fulfilled only when the total effective payload handled by the allocated coalition members satisfies the demand $R_m$. Given the strict time constraints, only the payload from AAVs arriving before the deadline $T_m^{end}$ is effective. The condition is expressed as

$$\sum_{n\in\text{Coal}_m}\mathbb{I}(t_m^{(n)}\le T_m^{end})\cdot a_m^{(n)}\ge R_m \tag{3}$$

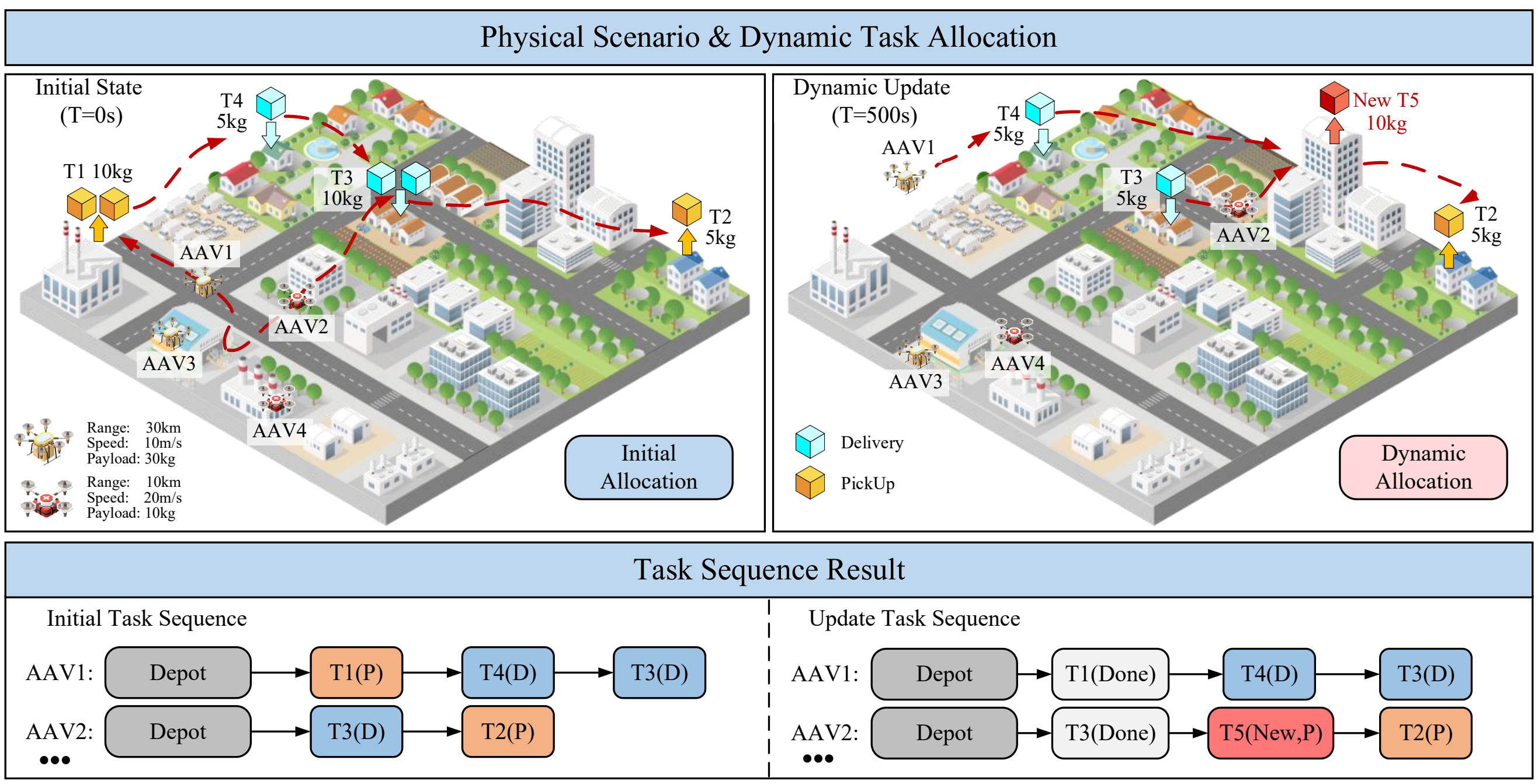

Fig.1 Illustration of the heterogeneous AAV logistics task allocation scenario. At T = 500 s, a newly emerged task triggers dynamic reallocation, leading the AAV swarm to update the task sequence to accommodate the new logistics demand. The newly emerged task T5 is highlighted in red.

where $\mathbb{I}(\cdot)$ is the indicator function, such that $\mathbb{I}(\cdot)=1$ if the condition holds, and $\mathbb{I}(\cdot)=0$, otherwise.

With these definitions, a global solution is characterized by the set of all payload allocation vectors $\left\{\mathcal{A}_1^{(*)}, \ldots, \mathcal{A}_M^{(*)}\right\}$. The corresponding overlapping coalition structure $\mathcal{OC}^{(*)}$ is induced by these allocation vectors through Equation (2).

The execution cost $C_m(\mathcal{A}_m)$ is formulated as a weighted sum of three performance metrics:

$$C_m(\mathcal{A}_m)=\omega_1 C_m^{load}(\mathcal{A}_m)+\omega_2 C_m^{time}(\mathcal{A}_m)+\omega_3 C_m^{op}(\mathcal{A}_m) \tag{4}$$

where $\omega_1,\omega_2,\omega_3$ are predesigned weight coefficients, $C_m^{load}(\mathcal{A}_m)$ is the payload deficiency penalty, $C_m^{time}(\mathcal{A}_m)$ is the service timeliness cost, and $C_m^{op}(\mathcal{A}_m)$ is the operational cost. By balancing service quality with resource consumption, this generalized logistics cost effectively quantifies the global optimality of logistics task allocation.

In Equation (4), $C_m^{load}(\mathcal{A}_m)$ quantifies the unmet payload requirement of task $m$. Inspired by the resource matching concept in [21], this penalty is defined as the complement of the valid payload fulfillment ratio, i.e.

$$C_m^{load}(\mathcal{A}_m)=\max\left\{0, 1-\sum_{n\in \mathrm{Coal}_m} \mathbb{I}(t_m^{(n)} \le T_m^{end}) \cdot a_m^{(n)} / R_m\right\} \tag{5}$$

where $C_m^{load}(\mathcal{A}_m)$ is lower bounded by zero, ensuring that no penalty is applied once the task requirement is fully satisfied.

$C_m^{time}(\mathcal{A}_m)$ is the service timeliness cost, evaluating how efficiently task $m$ is fulfilled within its time window. For a cooperatively executed task, $C_m^{time}(\mathcal{A}_m)$ continues to increase until the last AAV arrives.

$$C_m^{time}(\mathcal{A}_m)=\begin{cases} \dfrac{\max\limits_{n\in \mathrm{Coal}(\mathcal{A}_m)} t_m^{(n)} - T_m^{start}}{T^{Hor}}, & \max\limits_{n\in \mathrm{Coal}(\mathcal{A}_m)} t_m^{(n)} \le T_m^{end} \\ 1, & \text{otherwise} \end{cases} \tag{6}$$

where $T^{Hor}$ denotes the planning horizon for normalization. The arrival time $t_m^{(n)}$ depends on the cumulative flight time.

Let $\boldsymbol{W}_n=\left\{w_0^{(n)},\ldots,w_k^{(n)},\ldots w_{K_n}^{(n)}\right\}$ denote the execution sequence of AAV $n$, where $w_k^{(n)} \in \mathcal{M} \cup \mathcal{D}$ represents the $k$-th node (which can be a task or a depot). The arrival time $\tau_k^{(n)}$ at the $k$-th node is defined as

$$\tau_k^{(n)}=\max\left\{T_{w_k^{(n)}}^{start}, \tau_{k-1}^{(n)}+\frac{L(w_{k-1}^{(n)},w_k^{(n)})}{v_n}\right\} \tag{7}$$

To ensure kinematic feasibility, the differential flatness algorithm [22] is employed to calculate the path length $L(w_{k-1}^{(n)},w_k^{(n)})$ between consecutive nodes. Hence, if task $m$ corresponds to the node $w_k^{(n)}$ in the sequence, its arrival time is given by $t_m^{(n)}=\tau_k^{(n)}$.

$C_m^{op}(\mathcal{A}_m)$ represents the operational cost, integrating both energy consumption and the penalty for exceeding maximum flight range. It is defined as the sum of the individual costs:

$$C_m^{op}(\mathcal{A}_m)=\sum_{n\in \mathrm{Coal}_m} q_n(m) \tag{8}$$

For each AAV $n$, the individual costs $q_n(m)$ is a piecewise function. Let $D_{cum}^{(n)}$ be the cumulative distance flown after completing task $m$. The cost is formulated to heavily penalize any violation of the maximum flight range $D_n^{\max}$.

$$q_n(m)=\frac{1}{E_{base}}\begin{cases} \alpha_n L(w_{k-1}^{(n)},w_k^{(n)}), & D_{cum}^{(n)} \le D_n^{\max} \\ \alpha_n D_n^{\max}+\lambda_n, & \text{otherwise} \end{cases} \tag{9}$$

where $E_{base}=\max\limits_{i\in\mathcal{N}}(\alpha_i D_i^{\max}+\lambda_i)$ is the maximum economic loss, $\alpha_n$ is the energy rate, and $\lambda_n$ is the intrinsic value.

### *B. Physical Constraints*

Given the above definitions, the constraints of Problem 1 are given as follows.

Each AAV $n$ must ensure the instantaneous payload never exceeds the maximum capacity $B_n^{\max}$. Let $b_k^{(n)}$ denote the payload carried by AAV $n$ after executing the $k$-th node in its sequence. The recursive relationship is

$$b_k^{(n)} = \begin{cases} b_{k-1}^{(n)} - a_{ts_k^{(n)}}^{(n)}, & w_k^{(n)} \in \mathcal{M}_D \\ b_{k-1}^{(n)} + a_{ts_k^{(n)}}^{(n)}, & w_k^{(n)} \in \mathcal{M}_P \\ \min(B_n^{\max}, \Omega_k^{(n)}), & w_k^{(n)} \in \mathcal{D} \end{cases} \tag{10}$$

$$\text{s.t.} \quad b_k^{(n)} \in [0, B_n^{\max}]$$

where $\Omega_k^{(n)}$ represents the cumulative demand of all delivery tasks scheduled between the current depot and the next depot.

Although AAVs can replenish payload at depots, their energy is constrained. Instead of imposing a hard cutoff that might hinder the exploration of the solution space, the range limit is incorporated as a soft constraint in Equation (9). However, the mission must satisfy the closed-loop requirement.

$$w_0^{(n)} \in \mathcal{D}, w_{K_n}^{(n)} \in \mathcal{D}, \forall n \in \mathcal{N} \tag{11}$$

Based on the above analysis, one knows that Problem 1 is a typical NP-hard combinatorial optimization problem, characterized by strong coupling between task allocation and resource constraints. Conventional coalition formation game (CFG) models, which restrict AAVs to joining only one coalition at a time, struggle to efficiently explore the vast solution space. To address this limitation and enable more flexible resource allocation, the overlapping coalition formation (OCF) game is utilized to solve the problem in the next section.

## III. Overlapping Coalition Formation Game Based On Reinforcement Learning

The performance of conventional OCF games is constrained by myopic and rigid task selection strategies, which struggle to adapt to dynamic logistics environments. To overcome this limitation, we propose a transformer-based soft actor-critic guided overlapping coalition formation (TSAC-OCF) algorithm. This approach replaces rigid rules with forward-looking task selection policies derived from reinforcement learning, thereby enhancing the optimality of OCF games.

### *A. Decisional Bottleneck of Conventional OCF Games*

In OCF games, AAVs are modelled as rational players that establish stable coalitions driven by specific utility functions and cooperative criteria [10]. Within this framework, the optimality of allocation schemes is strictly dependent on the task selection strategies. However, existing works primarily employ random or heuristic rules, resulting in a decisional bottleneck that constrains global optimality [14], [15], [23].

#### *1) Random Selection*

Serving as a baseline, this strategy acts uniformly over the action space. AAVs select tasks with equal probability, neglecting critical constraints such as task urgency or spatial proximity. Consequently, this strategy suffers from severe inefficiency; as illustrated in Fig.2 (a), the blind selection process often leads AAVs to choose low-value tasks, failing to capture high-value ones.

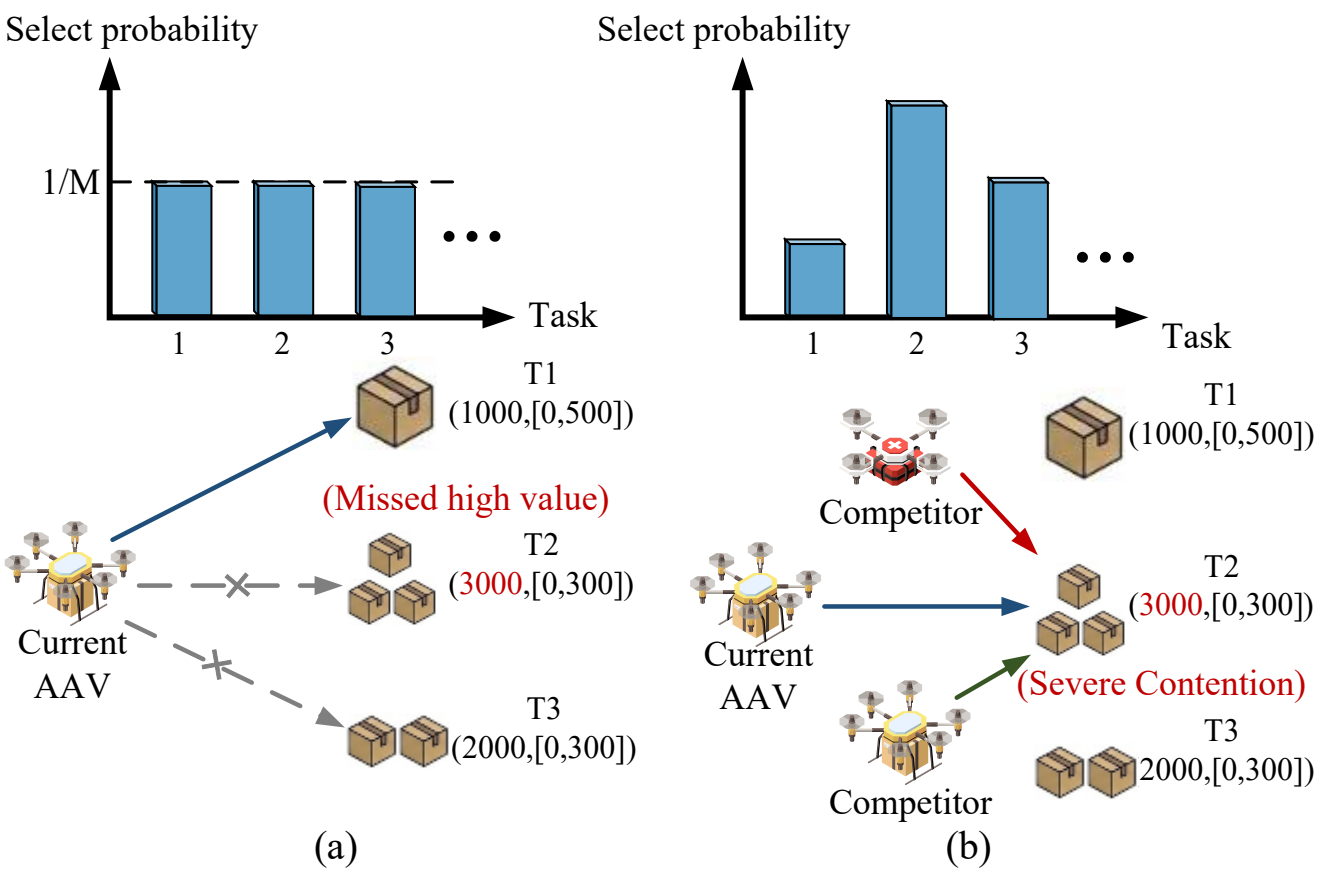

Fig.2. Conventional selection strategy in OCF games. (a) Random selection. (b) Heuristic-based selection.

#### *2) Heuristic-Based Selection*

To improve upon the random strategy, heuristic rules utilize predefined mathematical formulas to prioritize high-value tasks. While computationally efficient, this strategy suffers from three intrinsic limitations [4]:

(1) Temporal Myopia: Heuristics typically maximize immediate local utility without considering long-term system states. Such myopic decision-making frequently drives the system into local optima, where AAVs exhaust resources on current lower-value tasks, leaving insufficient capacity for subsequent high-priority orders.

(2) Resource Contention: Driven by homogeneous greedy logic, multiple AAVs simultaneously compete for the same high-value task. This leads to redundant resource allocation, where the aggregated payload for a specific task far exceeds its demand (e.g., Task 2 in Fig.2 (b)). Meanwhile, other tasks (e.g., Task 1 and Task 3) remain underserved.

(3) Static Rigidity: Heuristic effectiveness relies heavily on parameter tuning, which requires extensive domain expertise. Furthermore, these static weights lack adaptability, limiting the strategy's performance in dynamic or unforeseen scenarios.

To overcome these bottlenecks, reinforcement learning (RL) offers a more adaptive alternative. By evaluating current allocation decisions in terms of their long-term impact on subsequent tasks, RL enables AAVs to learn forward-looking task selection strategies, making it particularly appealing for OCF game-based logistics task allocation.

### *B. Dynamic Policy Learning via Transformer-based SAC*

In dynamic logistics scenarios, the stochastic emergence of time-sensitive tasks results in fluctuating input lengths, posing a structural challenge for these fixed-length models. To address this problem, a transformer-based soft actor-critic network is designed in this paper. By leveraging the multi-head self-attention mechanism to aggregate variable-scale logistics states [24], the proposed architecture inherently accommodates dynamic logistic task sequences and captures key spatio-temporal dependencies.

#### *1) Attention-Based Network Architecture*

To accommodate the time-variant property of logistics characterized by stochastically fluctuating task sequences, an attention-based network architecture is designed in Fig.3. The overall framework is logically structured into four functional modules, corresponding to the bold blue labels in the figure. The first three constitute the attention-based network architecture, while the fourth, policy evaluation, is detailed in the SAC training procedure in the next subsection.

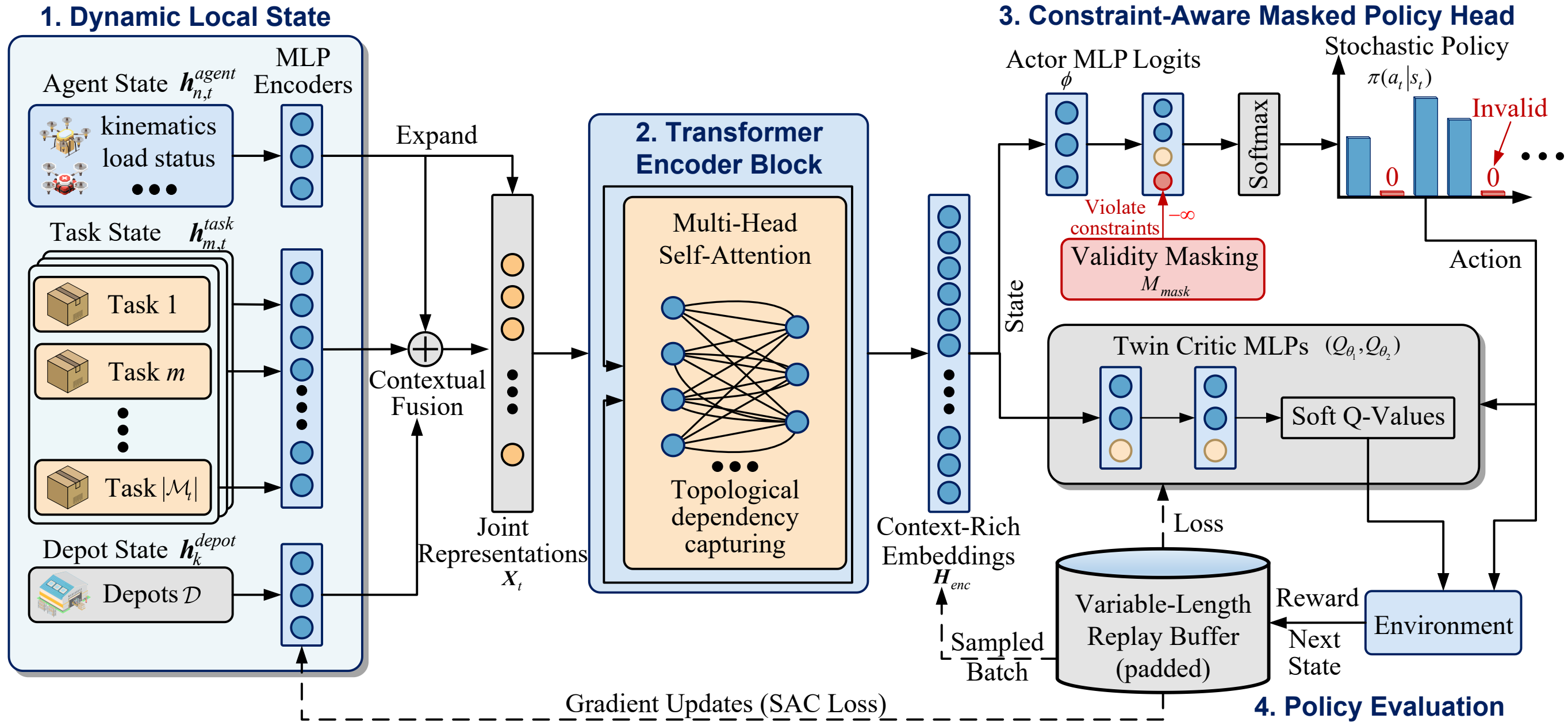

Fig.3. Constraint-aware Transformer-based SAC architecture for forward-looking task selection in dynamic logistics scenarios. The proposed architecture learns context-aware and physically feasible task-selection policies from dynamic logistics task sequences.

a) Dynamic State Formulation and Contextual Fusion

The local observation of AAV $n$ at time $t$ is formulated as a dynamic composite state $s_t = \{\boldsymbol{h}_{n,t}^{agent}, \{\boldsymbol{h}_{m,t}^{task}\}_{m\in\mathcal{M}_t}, \{\boldsymbol{h}_k^{depot}\}_{k\in\mathcal{D}}\}$, where $\mathcal{M}_t \subseteq \mathcal{M}$ is currently available tasks.

As denoted by the dynamic local state module in Fig.3, these logistics information are first projected into a unified high-dimensional feature space via separate multi-layer perceptron (MLP) encoders. Subsequently, to resolve the heterogeneity between the AAV and various candidate nodes, the encoded agent feature is expanded and concatenated with each encoded task and depot feature. This contextual fusion generates a set of joint representations $\boldsymbol{X}_t = \{\boldsymbol{x}_1, \ldots, \boldsymbol{x}_{|M_t|+|\mathcal{D}|}\}$, where each $\boldsymbol{x}_i$ couples the intrinsic attributes of AAV $n$ with the demands of a specific candidate target.

b) Transformer Encoder Block

To capture the complex dependencies among randomly distributed tasks, the joint representations $\boldsymbol{X}_t$ are fed into a transformer encoder block. By employing the multi-head self-attention (MHSA) mechanism, the encoder models global interactions across all node pairs in parallel. Without relying on a fixed input order, the MHSA mechanism captures global interactions among candidate nodes and supports flexible modeling of their relational structure [24]. The attention weights are computed as

$$\text{Attention}(\boldsymbol{Q},\boldsymbol{K},\boldsymbol{V}) = \text{Softmax}\left(\frac{\boldsymbol{Q}\boldsymbol{K}^T}{\sqrt{d_{\text{model}}}}\right)\boldsymbol{V} \qquad (12)$$

where $\boldsymbol{Q}$, $\boldsymbol{K}$, $\boldsymbol{V}$ denote the query, key, and value matrices projected from $\boldsymbol{X}_t$, and $d_{\text{model}}$ is the scaling dimension.

Rather than treating all inputs equally, this mechanism dynamically emphasizes candidate tasks that are more critical to logistics execution, such as urgent deliveries or spatially concentrated demands. Consequently, the variable-length sequence is aggregated into context-rich feature embeddings, denoted as $\boldsymbol{H}_{enc} \in \mathbb{R}^{(|M_t|\times|\mathcal{D}|)\times d_{\text{model}}}$.

c) Constraint-Aware Masked Policy Head

Corresponding to the upper-right module in Fig.3, upon extracting the context-rich embeddings $\boldsymbol{H}_{enc}$, the actor network maps the high-level features to a probability distribution over the discrete action space $\mathcal{M}_t \cup \mathcal{D}$. However, the standard Softmax function risks assigning non-zero probabilities to physically infeasible tasks, leading to execution failures [25].

To strictly enforce the hard constraints defined in Section II, a validity masking mechanism is integrated. Specifically, a mask vector $\boldsymbol{M}_{mask} \in \{0, -\infty\}^{M_t}$ is generated at each decision step, where tasks violating the maximum flight range constraints $D_n^{\max}$ or time window deadlines $T_m^{end}$ are assigned negative infinity. This mask is directly added to the pre-activation logits.

$$\pi_\phi(\alpha_t|s_t) = \text{Softmax}(\text{MLP}(\boldsymbol{H}_{enc}) + \boldsymbol{M}_{mask}) \qquad (13)$$

By restricting the probability mass solely to feasible candidates, this mechanism ensures that the selected task is physically executable under logistics constraints, while also improving learning efficiency by avoiding invalid exploration.

*2) Transformer-based SAC Training*

To learn optimal policies from the context-rich embeddings $\boldsymbol{H}_{enc}$, the soft actor-critic (SAC) algorithm [26] is employed for policy evaluation in the lower-right of Fig.3. Leveraging a maximum-entropy framework, SAC optimizes both the expected cumulative return and the policy entropy. This encour-ages the AAV to explore diverse feasible tasks during training and avoid myopic decisions in dynamic logistics environments.

a) Critic Network Optimization

The critic network estimates the soft action-value function $Q_{\theta_i}(s_t,a_t)$. To mitigate overestimation bias, we utilize a twin critic architecture parameterized by $\theta_1,\theta_2$. The parameters are updated by minimizing the soft Bellman residual.

$$J_Q(\theta_i)=\mathbb{E}_{(s_t,a_t,r_t,s_{t+1})\sim\mathcal{B}}\left[\left(Q_{\theta_i}(s_t,a_t)-y_t\right)^2\right] \tag{14}$$

Adapted for the discrete action space, the target value $y_t$ is computed over the actor's probability distribution, avoiding the stochastic sampling used in standard continuous SAC.

$$\begin{aligned} y_t = r_t + \gamma \mathbb{E}_{s_{t+1}\sim\mathcal{B}} \Bigg[ \sum_{a^{'}\in\mathcal{M}_{t+1}} \pi_\phi\left(a^{'}|s_{t+1}\right)\Big(\min_{j=1,2} Q_{\bar{\theta}_j}\left(s_{t+1},a^{'}\right) \\ - \alpha_{ent}\log\pi_\phi\left(a^{'}|s_{t+1}\right)\Big)\Bigg] \end{aligned} \tag{15}$$

where $\gamma\in(0,1)$ is the discount factor, $\alpha_{ent}$ is the temperature parameter regulating policy stochasticity, and $\mathcal{M}_{t+1}$ denotes the set of feasible tasks at the next time step, and $a^{'}$ is defined as a specific feasible action evaluated at state $s_{t+1}$.

b) Actor Network Optimization

The actor network, parameterized by $\phi$, is optimized by minimizing the Kullback-Leibler (KL) divergence between its policy distribution and the Boltzmann distribution induced by the soft Q-values [27]. For the discrete action space, the objective is defined as

$$\begin{aligned} J_\pi(\phi) = \mathbb{E}_{s_t\sim\mathcal{B}} \Bigg[ \sum_{a_t\in\mathcal{M}_t} \pi_\phi\left(a_t|s_t\right)\Big(\alpha_{ent}\log\pi_\phi\left(a_t|s_t\right) \\ - \min_{j=1,2} Q_{\theta_j}\left(s_t,a_t\right)\Big)\Bigg] \end{aligned} \tag{16}$$

where $\pi_\phi$ integrates the validity mask defined in Equation (13). This ensures probability mass is strictly confined to physically feasible tasks, effectively driving the AAV to select high-value targets while maintaining sufficient entropy for robust exploration.

c) Variable-Length Experience Replay

In dynamic logistics scenarios, stochastic order arrivals continuously change the set of available tasks, resulting in transitions with variable-length inputs. Since standard replay buffers are designed for fixed-dimensional experiences, a variable-length experience replay mechanism is developed to support training under such dynamic conditions.

Specifically, for a sampled mini-batch of $|\mathcal{B}|$ transitions, let $l_i$ denote the effective sequence length of the $i$-th transition. All sequences are padded to the maximum length $L_{\max}=\max_{i\in[1,N]} l_i$. To prevent these padding elements from corrupting the gradient updates, we construct a binary mask matrix $\boldsymbol{P}_{mask}\in\{0,1\}^{|\mathcal{B}|\times L_{\max}}$,where the entry $p_{i,j}$ is defined as 1 *if* $j\le l_i$ and 0 otherwise. During backpropagation, the loss function $\ell(\cdot)$ is modulated by this mask, ensuring gradients are derived solely from valid interactions.

$$\mathcal{L}_\theta=\frac{1}{\sum p_{i,j}}\sum_{i=1}^{|\mathcal{B}|}\sum_{j=1}^{L_{\max}} p_{i,j}\cdot\ell\left(s_{i,j},a_{i,j},\theta\right) \tag{17}$$

Ultimately, by integrating the transformer with SAC, the proposed framework enables the AAV to learn a context-aware policy that adapts to variable inputs and logistics constraints, thereby providing more flexible and forward-looking guidance for subsequent coalition formation.

### *C. RL-Guided Online OCF Mechanism*

Building upon the learned context-aware policy, this section describes how the policy is embedded into the OCF process to guide online coalition formation in dynamic logistics environments. The overall execution flow is illustrated in Fig. 4, and the implementation consists of the following three components.

*1) Asynchronous Iterative Framework*

To mitigate the resource contention in synchronous decision-making, the proposed RL-guided OCF mechanism adopts an asynchronous iterative framework. This algorithm decomposes the global coalition formation into sequential local optimization steps. At the beginning of each iteration $k$, a random permutation function is applied to generate a decision sequence $\mathcal{O}_k$.

$$\mathcal{O}_k=\langle n_1,n_2,\ldots,n_N\rangle, n_i\in\mathcal{N}, n_i\ne n_j \tag{18}$$

where $n_i$ represents the $i$-th activated AAV. This randomization prevents specific AAVs from consistently monopolizing high-value tasks, thereby ensuring fairness and avoiding sequence-dependent local optima.

When it is the turn of $n_i$, it observes the partial coalition structure that has already been updated by the preceding AAVs $\{n_1,\ldots,n_{i-1}\}$. To ensure information consistency, decisions are driven by the dynamic residual demand $\hat{R}_m$ rather than the static initial demand $R_m$. Derived from the valid payload fulfillment ratio defined in Equation (5), the residual demand $\hat{R}_m$ observed by $n_i$ is formulated as

$$\hat{R}_m(s_{k,i})=\max\left\{0, R_m-\sum_{j\in\mathcal{N}\setminus\{n_i\}}\mathbb{I}(t_m^{(j)}\le T_m^{end})\cdot a_m^{(j)}\right\} \tag{19}$$

Based on this dynamic observation, AAV $n_i$ leverages the pre-trained transformer policy to target unserved demands. This asynchronous execution immediately propagates state changes to the subsequent AAV $n_{i+1}$, effectively eliminating redundant allocations and multi-AAV conflicts.

*2) Context-Aware Policy Inference*

Within the asynchronous decision chain, AAV $n_i$ selects its target $m^*$ via the pre-trained transformer policy. First, it constructs the dynamic local state $s_{k,i}$ based on the feature definitions provided in Sec. III-B.

$$s_{k,i}=<\boldsymbol{h}_{n_i}^{agent},\{\boldsymbol{h}_m^{task}\}_{m\in\mathcal{M}},\{\boldsymbol{h}_d^{depot}\}_{d\in\mathcal{D}}> \tag{20}$$

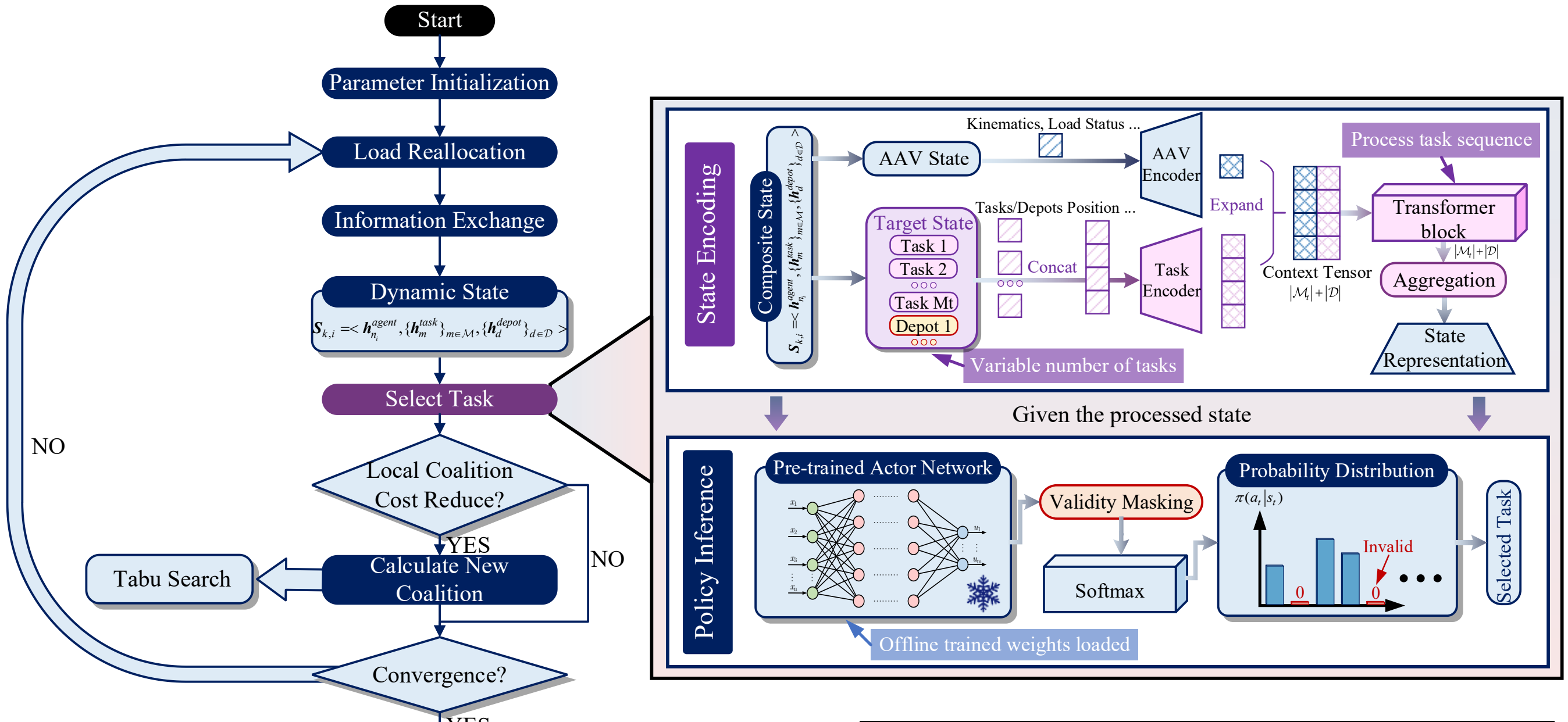

Fig. 4. Online execution flow of the proposed TSAC-OCF method.

where the agent embedding $\boldsymbol{h}_{n_i}^{agent}$ encodes the instantaneous kinematics and payload status of AAV $n_i$. Crucially, the task embeddings $\{\boldsymbol{h}_m^{task}\}_{m\in\mathcal{M}}$ are updated with the residual demand $\hat{R}_m$ and real-time coalition feedback.

This dynamic updating captures the precise spatiotemporal dependencies of iteration $k$. $s_{k,i}$ is then fed into the frozen policy network. To enforce physical constraints during online execution, the validity mask $\boldsymbol{M}_{\text{mask}}$ is applied to the raw logits $z(s_{k,i})\in\mathbb{R}^{|\mathcal{M}|}$. The executable probability distribution $\pi_\phi$ is derived as

$$\pi_\phi(m|s_{k,i})=\frac{\exp(z_m(s_{k,i})+\boldsymbol{M}_{\text{mask}}(m))}{\sum_{j\in\mathcal{M}}\exp(z_j(s_{k,i})+\boldsymbol{M}_{\text{mask}}(j))} \tag{21}$$

where $\boldsymbol{M}_{\text{mask}}(m)=0$ for feasible task $m$, and $-\infty$ otherwise. Finally, the target task is selected by stochastic sampling $m^*\sim\pi_\phi(\cdot|s_{k,i})$. Rather than deterministically exploiting the highest-probability target, this stochastic sampling ensures continuous exploration, effectively preventing the swarm from prematurely converging to sub-optimal coalitions.

*3) Hybrid Stabilization Mechanism*

While the context-aware policy proposes high-potential tasks, blindly accepting candidate updates induced by stochastic policy inference can lead to cyclic oscillations. To guarantee convergence to a Nash-stable solution, the proposed method incorporates a hybrid stabilization mechanism to regulate how candidate tasks proposed by the learned policy are applied to coalition updates. Specifically, after a task is selected, the mechanism determines one of two coalition-update operations according to whether the task has already appeared in the current execution sequence.

If the selected task has not appeared in the execution sequence $m^*\notin\boldsymbol{W}_n$, an ADD operation is performed, through which the AAV attempts to join the coalition associated with that task. Otherwise, a REMOVE operation is performed,

**Algorithm 1**. TSAC-OCF

**Input** $\mathcal{N}$, $\mathcal{M}$, $K_{\max}$, $C_{\max}$

**Output** Final resource allocation structure $\mathcal{OC}^{(*)}$

1: **Initialization:** $k=0$, $c=0$, $\mathcal{OC}^{(k)}$, $\mathcal{T}=\varnothing$
2: **loop** $\forall n\in\mathcal{N}$
3: AAV $n$ obtains current information, then builds $s_{k,i}$
4: Input $s_{k,i}$ to trained model to obtain the desired task $m^*$
5: Calculate the new resource allocation structure $\mathcal{OC}_{new}^{(k)}$
6: **if** $\mathcal{OC}_{new}^{(k)}\succ_n\mathcal{OC}^{(k)}$ **then** $\mathcal{OC}^{(k+1)}=\mathcal{OC}_{new}^{(k)}$ $c=0$
7: **else** $\mathcal{OC}^{(k+1)}=\mathcal{OC}^{(k)}$ $c=c+1$
8: **end if**
9: $k=k+1$, $\mathcal{OC}^{(k)}\to\mathcal{T}$
10: **end loop if** $k>K_{\max}$ or $c>C_{\max}$

through which the AAV attempts to release its current allocation on the selected task and update the associated coalition arrangement accordingly. To drive the coalition formation process toward a stable solution, both operations are accepted if and only if the resulting coalition structure strictly reduces the coalition cost, i.e.,

$$C(\mathcal{OC}_{new})<C(\mathcal{OC}_{current}) \tag{22}$$

where $\mathcal{OC}_{current}$ is the current coalition structure and $\mathcal{OC}_{new}$ is the new coalition structure.

To prevent infinite deadlocks (e.g., an AAV repeatedly adding and removing the same task), a tabu list $\mathcal{T}$ is introduced to record the existing coalition structures [28]. Any action that results in coalition structures already recorded in $\mathcal{T}$ is prohibited. The pseudocode for TSAC-OCF is presented as follows.

By replacing conventional heuristics with a forward-looking policy learned via reinforcement learning, TSAC-OCF empowers AAVs to form more efficient coalition structures, enhancing the global optimality of multi-AAV task allocation.

## D. Coalition Structures Stability Analysis

To theoretically guarantee the convergence of TSAC-OCF, the concepts of an exact potential game (EPG) are introduced.

***Definition 1 (Exact potential game*** [29]***).*** A game is an EPG

if and only if there exists a potential function $\Phi$ such that:

$$C_n(\chi_n^{'},\chi_{-n})-C_n(\chi_n,\chi_{-n})=\Phi(\chi_n^{'},\chi_{-n})-\Phi(\chi_n,\chi_{-n}) \quad (23)$$

where $\chi_n$ and $\chi_n^{'}$ represents the strategy of AAV $n$ before and after performing the switching criterion, and $\chi_{-n}$ is the strategy of all AAVs except $n$ .

Based on the definition, we present the following theorem.

***Theorem 1:*** The OCF game under cooperative criteria possesses at least one stable coalition structure. TSAC-OCF is guaranteed to converge to a stable coalition structure in finite iterations, regardless of the initial coalition structure.

***Proof:*** Consider a scenario where AAV $n$ unilaterally changes its strategy from $\chi_n$ to $\chi_n^{'}$, while the strategies of all other AAVs, denoted by $\chi_{-n}$, remain fixed. The difference in its individual cost function is expressed as

$$\begin{aligned} C_n\left(\chi_n^{'},\chi_{-n}\right)-C_n\left(\chi_n,\chi_{-n}\right)&=c_n\left(\chi_n^{'},\chi_{-n}\right)-c_n\left(\chi_n,\chi_{-n}\right)\\ &+\sum_{i\in Mem\left(\mathcal{A}^I\right)\setminus\{n\}}\left[c_i\left(\chi_i^{'},\chi_{-i}\right)-c_i\left(\chi_i,\chi_{-i}\right)\right]\\ &+\sum_{i\in Mem\left(\mathcal{A}^J\right)\setminus\{n\}}\left[c_i\left(\chi_i^{'},\chi_{-i}\right)-c_i\left(\chi_i,\chi_{-i}\right)\right]\\ &+\sum_{\mathcal{A}^K\in\boldsymbol{SC}}\sum_{i\in Mem\left(\mathcal{A}^K\right)}\left[c_i\left(\chi_i^{'},\chi_{-i}\right)-c_i\left(\chi_i,\chi_{-i}\right)\right] \end{aligned} \quad (24)$$

After defining $C_n$, we construct the potential function $\Phi$:

$$\Phi\left(\chi_n,\chi_{-n}\right)=\sum_{n\in\mathcal{N}}c_n\left(\chi_n,\chi_{-n}\right) \quad (25)$$

According to Equation (25) and substituting the definition of $c_n$:

$$\begin{aligned} \Phi\left(\chi_n^{'},\chi_{-n}\right)-\Phi\left(\chi_n,\chi_{-n}\right)&=c_n\left(\chi_n^{'},\chi_{-n}\right)-c_n\left(\chi_n,\chi_{-n}\right)\\ &+\sum_{i\in Mem\left(\mathcal{A}^I\right)\setminus\{n\}}\left[c_i\left(\chi_i^{'},\chi_{-i}\right)-c_i\left(\chi_i,\chi_{-i}\right)\right]\\ &+\sum_{i\in Mem\left(\mathcal{A}^J\right)\setminus\{n\}}\left[c_i\left(\chi_i^{'},\chi_{-i}\right)-c_i\left(\chi_i,\chi_{-i}\right)\right]\\ &+\sum_{\mathcal{A}^K\in\boldsymbol{SC}}\sum_{i\in Mem\left(\mathcal{A}^K\right)}\left[c_i\left(\chi_i^{'},\chi_{-i}\right)-c_i\left(\chi_i,\chi_{-i}\right)\right] \end{aligned} \quad (26)$$

Combining Equation (24) and Equation (26):

$$\Phi(\chi_n^{'},\chi_{-n})-\Phi(\chi_n,\chi_{-n})=C_n(\chi_n^{'},\chi_{-n})-C_n(\chi_n,\chi_{-n}) \quad (27)$$

Equation (27) demonstrates that any unilateral decrease in an AAV's cost results in an identical decrease in the global potential function. According to the definition of the EPG, the existence of at least one pure-strategy NE is guaranteed.

Furthermore, given the discrete property of resources carried by AAVs and demanded by tasks, the state space of all possible coalition structures is finite. According to the finite improvement property (FIP) [29] of potential games, any sequence of strategy changes that strictly decreases a potential function over a finite state space must converge to a fixed point. Therefore, TSAC-OCF can converge to a stable NE point within a finite number of iterations. **Theorem 1** is proved.

### *E. Complexity Analysis*

To assess the computational feasibility of TSAC-OCF for real-world scenarios, we analyze its worst-case time complexity. The computational overhead per iteration comprises three main phases:

(1) State Encoding: The transformer encoder processes a sequence of $M$ tasks. Dominated by the self-attention mechanism, the feature extraction incurs a complexity of $O\left(M^2\right)$.

(2) Action Selection: Generating the probability distribution and applying the validity mask requires $O(M)$. The subsequent probability sorting for task selection takes $O(M\log M)$.

(3) Tabu Search: Checking the newly formed coalition against the Tabu list of length $L$ requires $O(L)$.

Given a maximum of $K_{\max}$ iterations, the total complexity is bounded by

$$C_{TSAC-OCF}=K_{\max}\left(O\left(M^2+M\log M+L\right)\right) \quad (28)$$

In practical urban logistics, the task scale $M$ and Tabu list length $L$ are bounded. Therefore, the dominant quadratic complexity $O(M^2)$ remains highly scalable, ensuring that TSAC-OCF strictly satisfies the real-time decision-making requirements of the AAV swarm.

## IV. Numerical Simulations

To evaluate the performance of the proposed TSAC-OCF algorithm, several typical scenarios for heterogeneous AAV logistics delivery are established. The algorithm is implemented in C++17 and executed on a workstation equipped with AMD Ryzen 9 7945HX 2.50 GHz CPU and 32 GB RAM.

### *A. Simulation Setup*

The operational environment is a 10 km×10 km area. To reflect diverse urban logistics scenarios, we consider two distinct AAV types with varying payload capacities and flight constraints: Heavy-duty AAVs (Type 1) and High-speed AAVs (Type 2). Their parameters are shown in TABLE 1.

TABLE 1 Parameters of heterogeneous AAVs

| Parameter | Type 1 | Type 2 |
|---|---|---|
| flight speed $v_n$ | 10 m/s | 20 m/s |
| maximum payload capacity $B_n^{\max}$ | 30 kg | 10 kg |
| maximum flight range $D_n^{\max}$ | 60 km | 30 km |
| energy rate $\alpha_n$ | 20 km | 10 km |
| intrinsic value $\lambda_n$ | 2000 | 500 |

Without loss of generality, the positions of the AAVs and tasks are randomly distributed within the area. The demands of each task are randomly assigned within predefined ranges. For all scenarios, the total mission planning horizon $T^{Hor}$ is set to 3600 s, the weight coefficients $(\omega_1,\omega_2,\omega_3)$ are set as $(100,10,1)$ to prioritize task completion.

### *B. Performance Analysis*

In this section, the performance of TSAC-OCF is comprehensively evaluated through two complementary comparisons. Horizontally, TSAC-OCF is benchmarked against distributed genetic algorithm (DGA) [30] and contract net protocol (CNP) [9] to demonstrate its superiority over traditional scheduling paradigms. Vertically, we compare TSAC-OCF with random overlapping coalition formation (ROCF) game [31] and Heuristic overlapping coalition formation (HOCF) game [14] to quantify the performance gains contributed by the RL-guided strategy within the OCF game.

*1) Mechanism Verification in a Small-Scale Scenario*

To validate the dynamic reallocation mechanism and spatiotemporal adaptability of TSAC-OCF, a small-scale scenario involving 4 heterogeneous AAVs, 2 depots, and 10 logistics tasks is established, with detailed configurations summarized in TABLE 2 and TABLE 3. To emulate a dynamic logistics environment, a subset of tasks emerges during flight, thereby triggering real-time reallocation. In TABLE 3, 'P' and 'D' denote pickup and delivery tasks, respectively, and 'TW' denotes the corresponding time window.

TABLE 2 Configurations of AAVs and Depots

| ID | Type | Position/m |
|---|---|---|
| $A_0$ | 1 | [-3712, -3970] |
| $A_1$ | 2 | [-3312, -3570] |
| $A_2$ | 2 | [3807, 3496] |
| $A_3$ | 2 | [3407, 3896] |
| $D_0$ | - | [-3500, -3500] |
| $D_1$ | - | [3500, 3500] |

As shown in Fig. 5 (a), TSAC-OCF exhibits rapid dynamic responsiveness. At the initial stage ($t$ = 0s), an original task sequence is generated for tasks $T_0 - T_3$, while **Event1** and **Event2** occur, the task sequences of all AAVs are adaptively reconfigured to incorporate the new orders ( $T_4 - T_6$ and $T_7 - T_9$ ). A distinctive feature of TSAC-OCF is the formation of overlapping coalitions for complex tasks. For instance, as highlighted by the purple dashed boxes in Fig. 5 (a), task $T_7$ is cooperatively executed by all 4 AAVs, ensuring that payload-heavy or time-sensitive tasks are prioritized.

The efficiency of this re-allocation is reflected in the total cost curve (lower panel of Fig. 5 (a)). Although the appearance of new tasks causes transient cost spikes at 25s and 90s, TSAC-OCF quickly converges to a new, lower-cost steady state, indicating that the re-allocation is timely.

TABLE 3 Configurations of tasks

| ID | Position/m | Type | Req/kg | TW/s |
|---|---|---|---|---|
| $T_0$ | [1377, -4730] | P | 10 | [0, 60] |
| $T_1$ | [2040, -1049] | D | 5 | [0, 60] |
| $T_2$ | [3905, 2040] | D | 5 | [0, 30] |
| $T_3$ | [1983, 1486] | P | 20 | [0, 60] |
| $T_4$ | [-3842, 529] | D | 40 | [25, 95] |
| $T_5$ | [-3727, 2747] | D | 10 | [25, 105] |
| $T_6$ | [3075, -974] | D | 10 | [25, 105] |
| $T_7$ | [-1256, -339] | P | 60 | [90, 140] |
| $T_8$ | [3075, -974] | D | 10 | [90, 190] |
| $T_9$ | [443, 3733] | P | 20 | [90, 220] |

To visualize this real-time decision-making, Fig. 5 (b) and Fig. 5 (c) provide spatial snapshots at $t$ = 90 s. Before reallocation, AAVs follow their original paths for tasks $T_4 - T_6$. After the new tasks are released, TSAC-OCF re-routes the swarm toward the newly emerged task $T_7$. This task requires a 60 kg pickup within the time window [90, 140], whereas the 4 AAVs provide capacities of 30, 10, 10, and 10 kg, respectively. Consequently, all 4 AAVs are redirected to $T_7$ to satisfy its payload demand within the deadline through a coordinated overlapping coalition. This comparison confirms that TSAC-OCF can promptly reorganize the swarm and adapt its coalition structure to the spatiotemporal requirements of newly arrived logistics tasks.

*2) Performance Superiority in a Large-Scale Scenario*

To evaluate the performance of TSAC-OCF in a large-scale logistics scenario, we expanded the problem scale to 32 heterogeneous AAVs and 80 dynamic tasks. As illustrated in the upper panel of Fig. 6, TSAC-OCF achieves the lowest cost (82.09), yielding a 33.92% reduction compared to the best-performing baseline, HOCF (124.22).

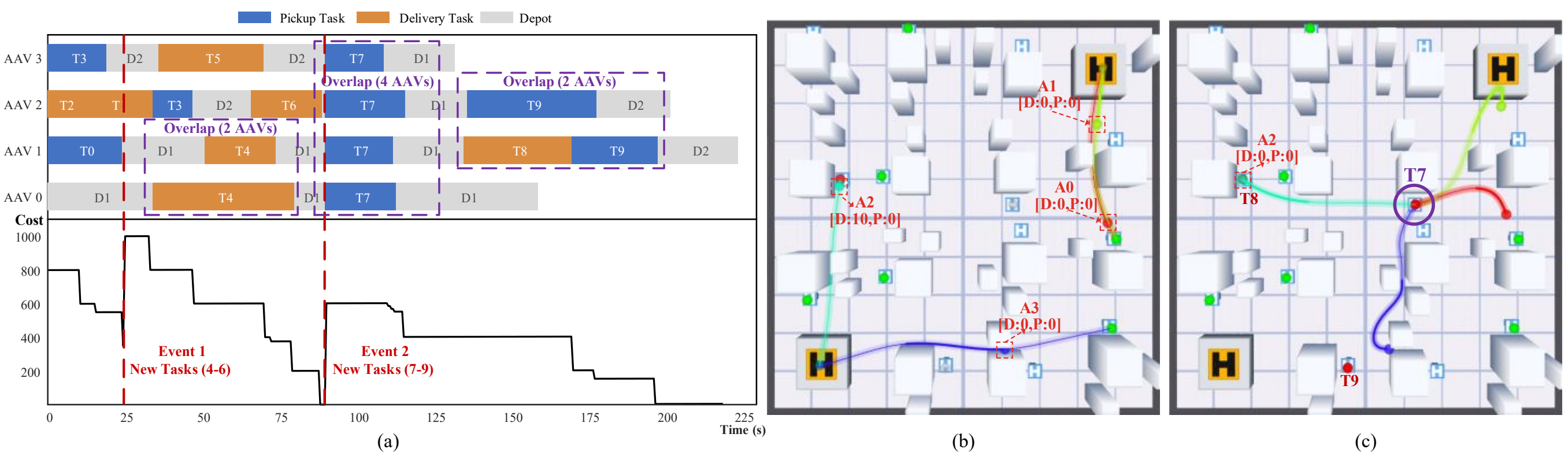

Fig. 5. Simulation results for the small-scale logistics scenario. (a) Logistics task Gantt chart and cost evolution during dynamic reallocation. (b) Spatial snapshot before reallocation at T = 90 s. (c) Spatial snapshot after reallocation at T = 90 s. The label [D:x, P:y] indicates the instantaneous onboard payload state of an AAV, where D and P represent the carried delivery load and pickup load, respectively.

The lower panel of Fig. 6 further explains this performance through coalition structure analysis. For tasks with a 10 kg payload requirement, both HOCF and TSAC-OCF tend to assign single light AAVs. However, as the payload demand increases to 30 kg or 60 kg, HOCF frequently expands coalition sizes, indicating a stronger reliance on aggregating multiple light AAVs. In contrast, TSAC-OCF maintains smaller coalition sizes for many heavy tasks, suggesting more efficient utilization of high-capacity AAV resources. This more compact coalition pattern reduces redundant resource aggregation and contributes to the lower total cost of TSAC-OCF, thereby demonstrating its advantage in logistics task allocation.

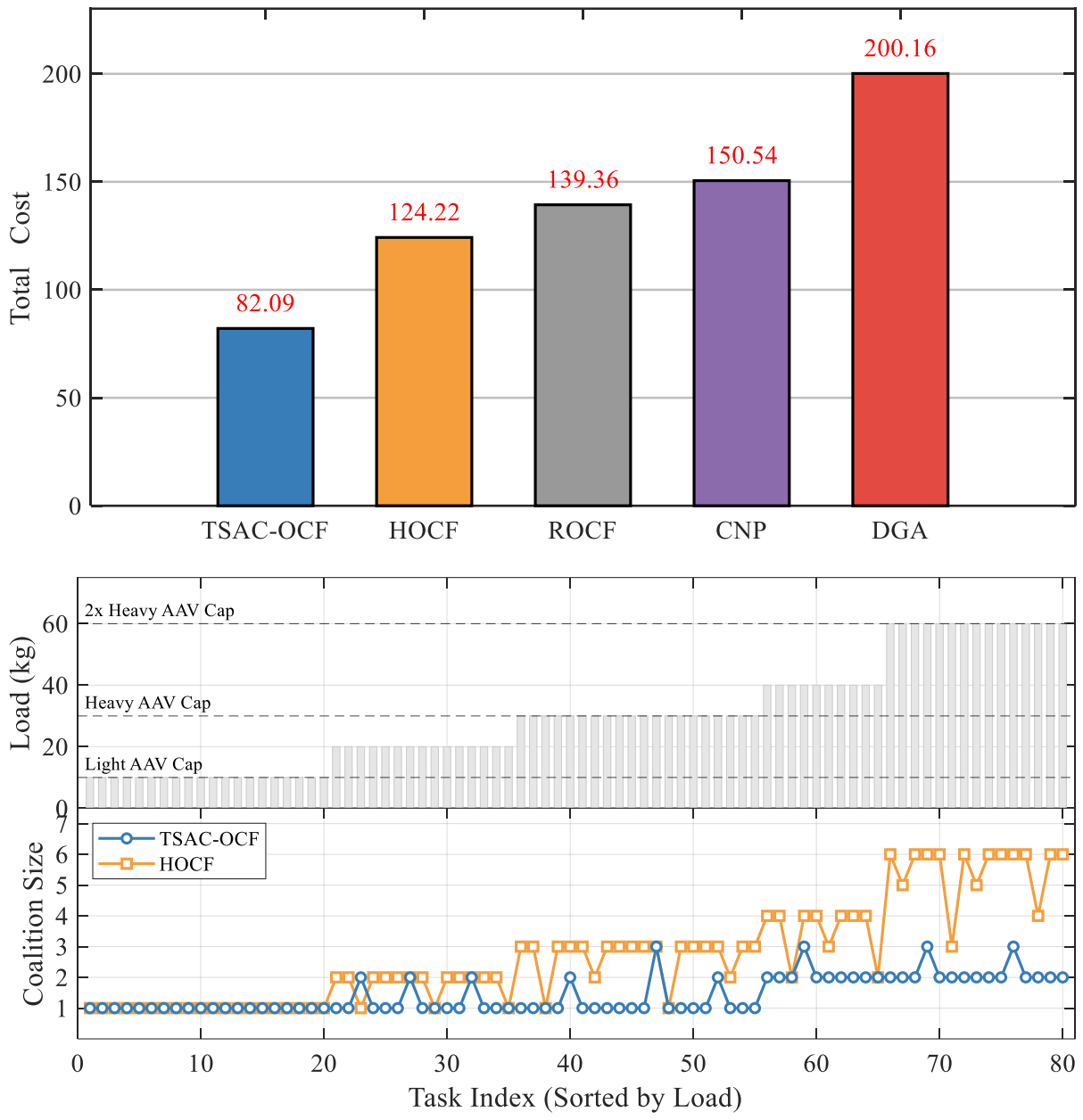

Fig. 6. Simulation results for the large-scale logistics scenario.

### 3) Statistical Reliability in Monte Carlo Simulations

To mitigate the bias caused by stochastic mission distributions and verify the statistical robustness of the proposed algorithm, 1000 independent Monte Carlo simulations are conducted across 4 scenario scales, namely 4/10, 8/20, 16/40, and 32/80. The boxplot results in Fig. 7 compare the total costs of 5 methods under these randomly generated logistics scenarios.

As shown in Fig. 7, CNP and DGA consistently yield the highest total costs across all scenario scales, reflecting their limited flexibility in handling overlapping resource coordination. In contrast, all OCF-based methods achieve lower costs, highlighting the advantage of coalition-based cooperation in dynamic logistics task allocation. Specifically, ROCF outperforms CNP by an average of 8.67% in total cost reduction.

Among the OCF-based methods, TSAC-OCF consistently attains the lowest median cost and maintains relatively narrow interquartile ranges across different scenario scales. This advantage becomes increasingly pronounced as the problem scale expands. In the 32/80 scenario, TSAC-OCF achieves a 39.76% reduction in average total cost compared with the second-best baseline, HOCF, demonstrating the benefit of replacing heuristic rules with the learned coalition-update policy.

In addition to lower average costs, TSAC-OCF exhibits fewer extreme outliers and more concentrated cost distributions. These results indicate that the proposed method is more robust to variations in task distributions and initial AAV positions. Overall, the multi-scale Monte Carlo analysis validates the superior performance and statistical robustness of TSAC-OCF across dynamic logistics scenarios of different scales.

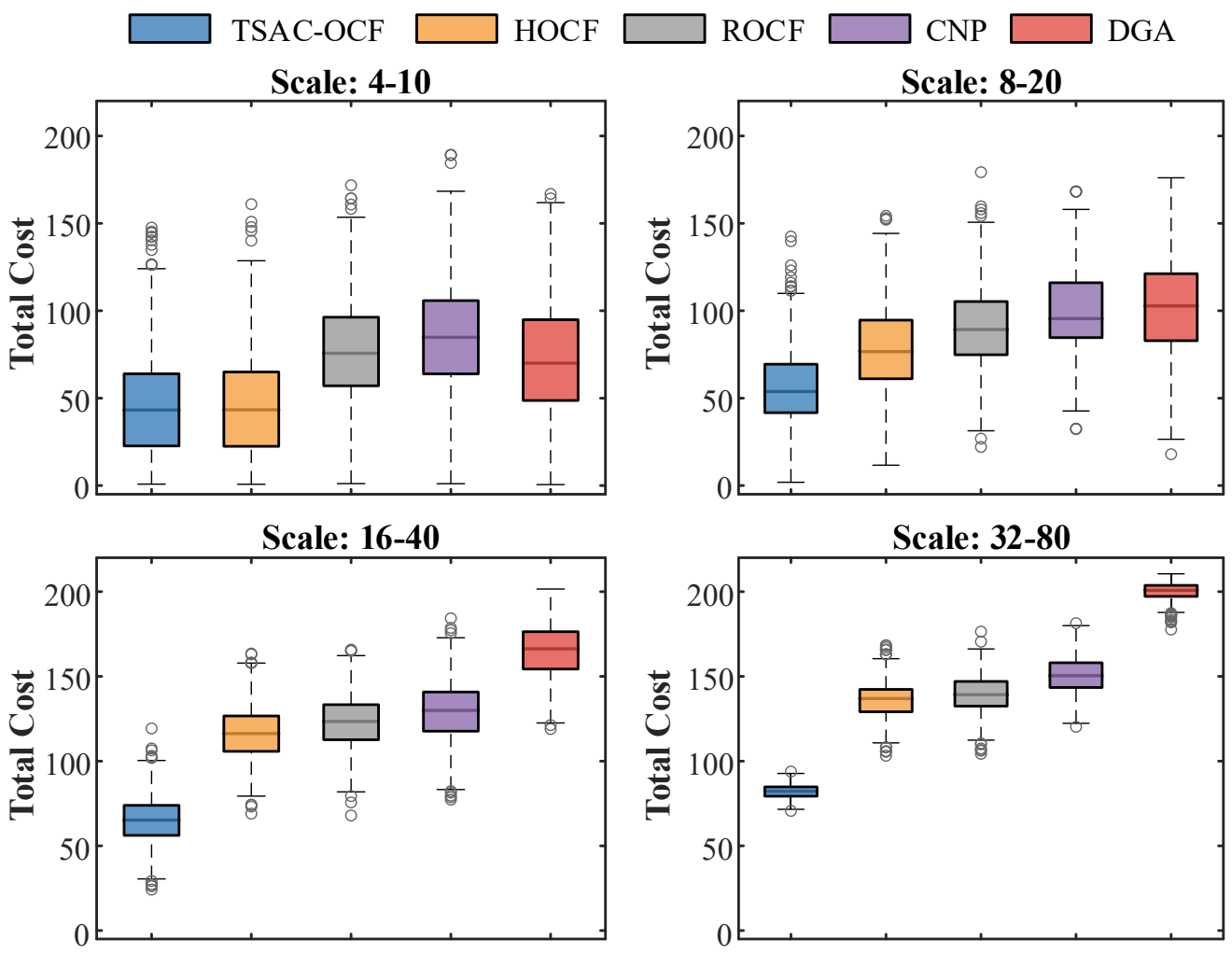

Fig. 7. Monte Carlo simulation results under different logistics scenario scales.

## V. Indoor Flight Experiments

To further bridge the gap between simulation and real-world application, indoor flight experiments are conducted to verify the practicality of TSAC-OCF. The experiments are conducted utilizing 4 Crazyflie 2.1 AAVs in a 4 m × 4 m × 2 m indoor testbed. Although the platforms are physically identical, heterogeneity is approximated by assigning different motion parameters, including flight speed and minimum turning radius, to emulate mobility differences within the swarm. Task-level heterogeneity is further reflected through differentiated execution and cooperation behaviors during collaborative logistics missions.

### A. Experimental Configurations

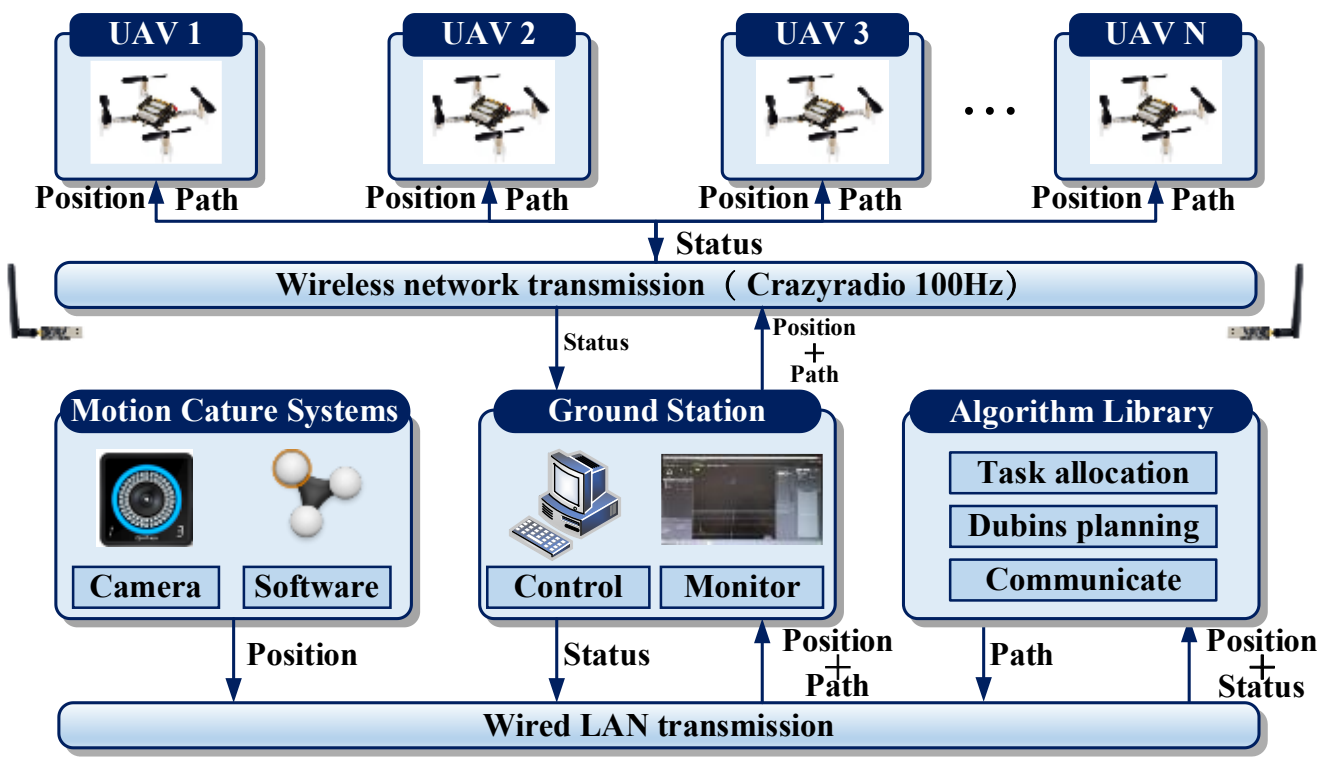

Fig. 8. The framework of indoor flight experiments.

Fig. 8 shows the framework of indoor flight experiments. The Optitrack motion capture systems acquire the real-time position of AAVs, transmitting the data to the ground station over a Wired LAN. Concurrently, AAVs send their status back

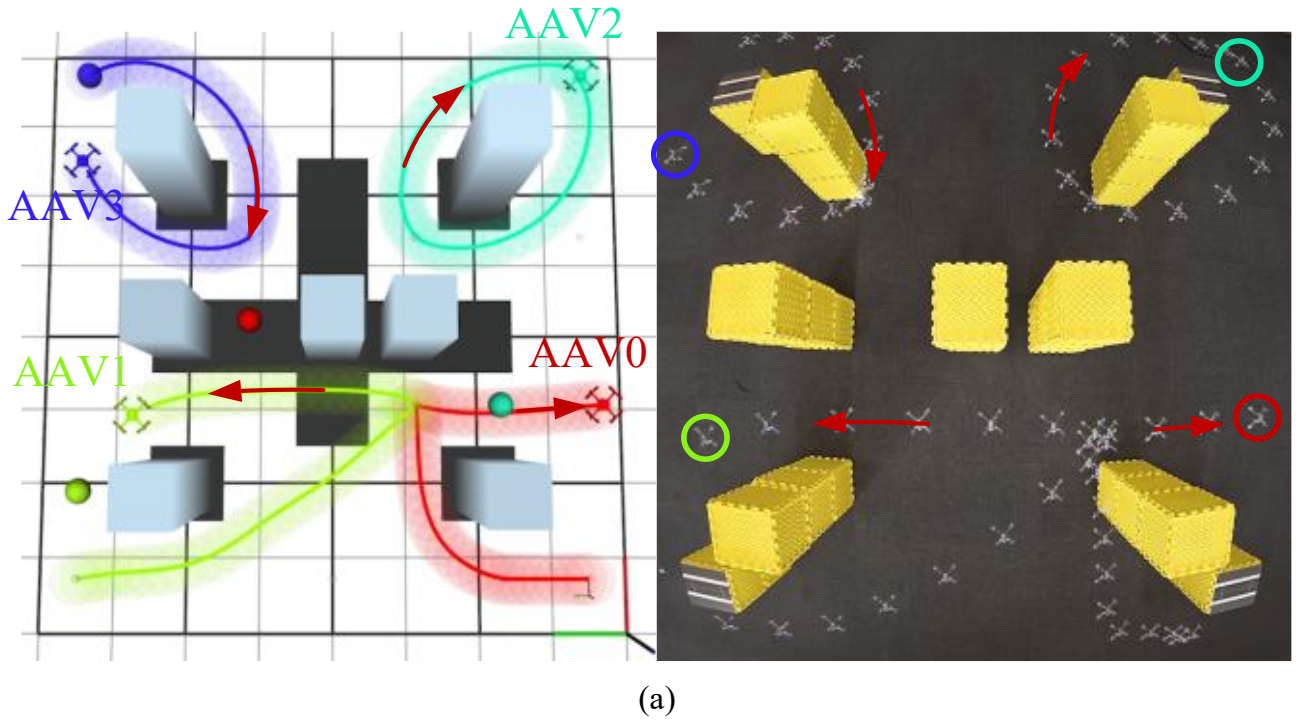

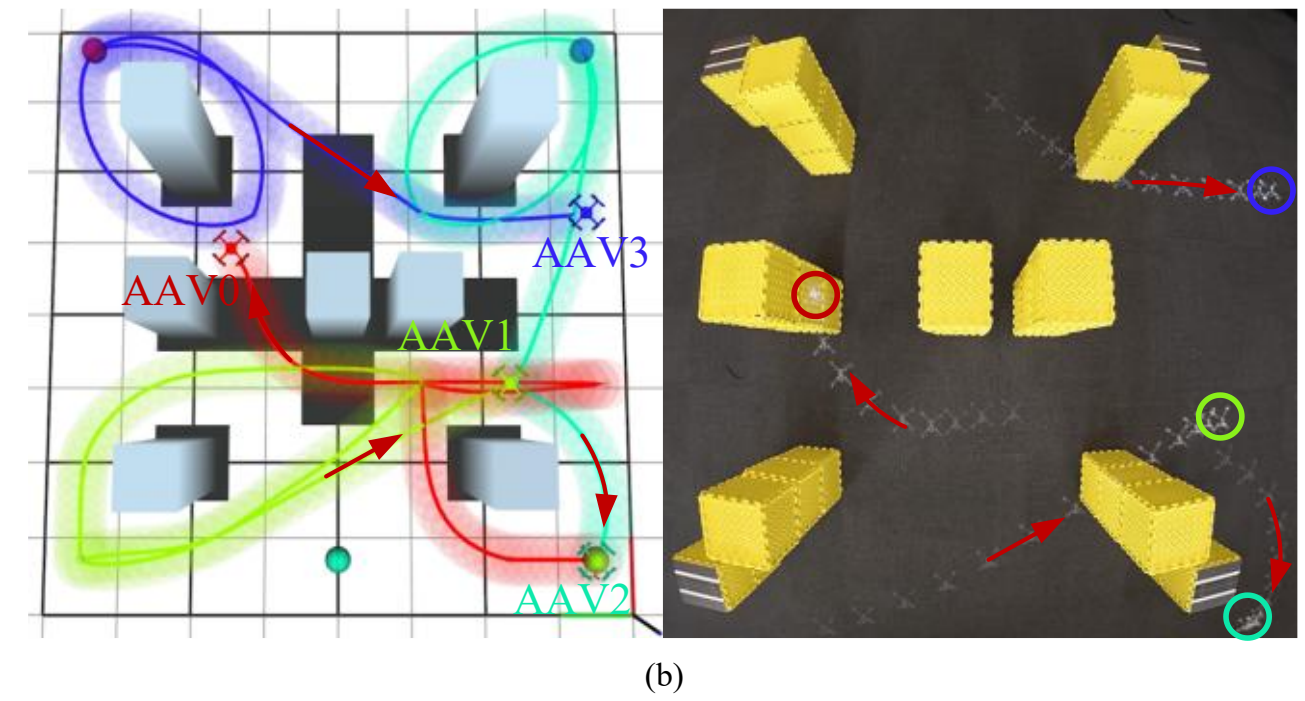

(a) (b)

Fig. 9. Indoor flight experiments for dynamic task reallocation. (a) First reallocation triggered by newly emerged tasks at T = 5 s. (b) Second reallocation triggered by further task arrivals at T = 10 s. (The detailed flight processes are shown in the supplementary video)

via a Crazyradio. The ground station aggregates and relays the data to the algorithm library for flight path generation. The calculated paths are then transmitted to the respective AAVs through a Crazyradio for task execution.

TSAC-OCF is embedded into the ground station to allocate tasks. After the algorithm generates respective task sequences for AAVs, the differential flatness trajectory planning is used to execute tasks. The flight paths are transmitted to each AAV and tracked by using autopilots.

### *B. Experimental Results*

As the experiments are conducted in a constrained indoor environment, the flight trajectories are proportionally mapped to the testbed while maintaining a minimum safety separation of 10 cm between AAVs.

The experimental results are shown in Fig. 9. Initially, the AAVs execute their assigned tasks based on the original allocation scheme. As depicted in Fig. 9 (a), the arrival of new tasks at $t$ = 5 s triggers an online reallocation, after which the flight trajectories are reconfigured to reflect the updated task assignments. As shown in Fig. 9 (b), another set of newly arrived tasks appears at $t$ = 10 s, leading to a second round of online reallocation during ongoing execution. Across these consecutive task-arrival events, the AAV swarm does not simply follow its preplanned paths; instead, it updates its trajectories online to accommodate newly emerged demands while preserving safe operation in the indoor testbed. These results demonstrate the practical feasibility of TSAC-OCF for dynamic task reallocation in real-time execution.

## VI. Conclusions

This paper proposed a transformer-based soft actor-critic guided overlapping coalition formation algorithm, termed TSAC-OCF, for dynamic task allocation in heterogeneous AAV logistics. By replacing heuristic rules with a learned task-selection policy, the proposed method improves coalition formation quality under dynamically arriving logistics demands. The transformer-based architecture enables the policy to process variable-length task sets and capture the spatiotemporal coupling among tasks, while the exact potential game analysis guarantees convergence of the resulting coalition formation process to a Nash-stable equilibrium within a finite number of iterations.

Numerical simulations and indoor flight experiments verified the effectiveness and practical feasibility of TSAC-OCF. In particular, the proposed method consistently achieved lower generalized logistics cost than the baseline methods, and its advantage became more pronounced as the problem scale increased. In the 32/80 scenario, TSAC-OCF achieved a 39.76% reduction in average total cost compared with HOCF.

The current model assumes reliable communication under an asynchronous iterative framework. Future work will investigate the robustness and effectiveness of the proposed algorithm in non-ideal communication environments.